\documentclass[12pt]{article}

\usepackage[utf8]{inputenc} 
\usepackage[T1]{fontenc}    
\usepackage{hyperref}       
\usepackage{url}            
\usepackage{booktabs}       
\usepackage{amsfonts}       
\usepackage{nicefrac}       
\usepackage{microtype}      
\usepackage{lipsum}
\usepackage{fancyhdr}       
\usepackage{graphics}
\usepackage{graphicx}       
\usepackage{multirow}
\usepackage{xcolor}
\usepackage{tablefootnote}
\usepackage[square,numbers]{natbib} 
\usepackage{arydshln}
\usepackage{adjustbox}
\usepackage{bm}
\usepackage{anyfontsize}
\usepackage{import}
\usepackage{placeins}
\usepackage{listings}
\usepackage{amsmath} 

\NewDocumentCommand\emojihuggingface{}{
    \includegraphics[scale=0.02]{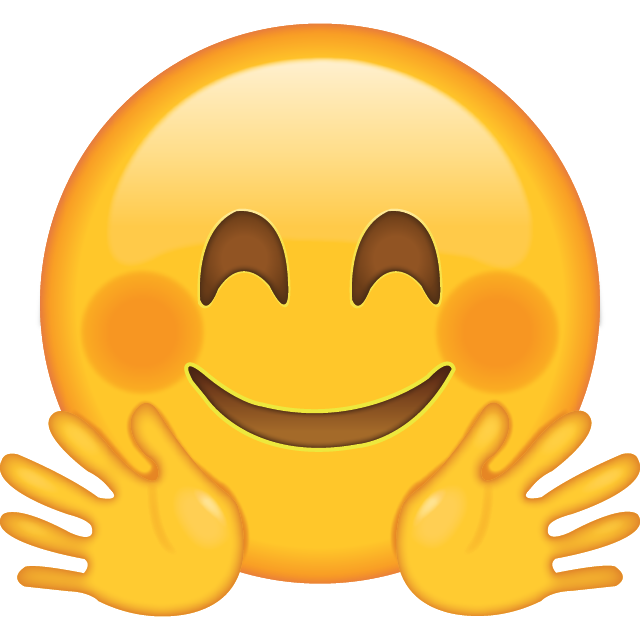}
}

\usepackage{pifont}
\newcommand{\myref}[1]{\href{#1}{#1}}

\title{TurkishBERTweet: Fast and Reliable Large Language Model for Social Media Analysis}

\author{
  Ali Najafi\textsuperscript{1},
  Onur Varol\textsuperscript{1,2,*}
}

\date{
    \begin{small}
    \textsuperscript{1}Faculty of Engineering and Natural Sciences, Sabanci University \\ \textsuperscript{2}Center of Excellence in Data Analytics, Sabanci University \\
    \textsuperscript{*}Corresponding author: \texttt{onur.varol@sabanciuniv.edu}
    \end{small}
}

\begin{document}
\maketitle

\begin{abstract} 
Turkish is one of the most popular languages in the world. Wide us of this language on social media platforms such as Twitter, Instagram, or Tiktok and strategic position of the country in the world politics makes it appealing for the social network researchers and industry. To address this need, we introduce \texttt{TurkishBERTweet}, the first large scale pre-trained language model for Turkish social media built using almost 900 million tweets. 
The model shares the same architecture as base BERT model with smaller input length, making \texttt{TurkishBERTweet} lighter than BERTurk and can have significantly lower inference time. 
We trained our model using the same approach for RoBERTa model and evaluated on two text classification tasks: Sentiment Classification and Hate Speech Detection. We demonstrate that \texttt{TurkishBERTweet} outperforms the other available alternatives on generalizability and its lower inference time gives significant advantage to process large-scale datasets. 
We also compared our models with the commercial OpenAI solutions in terms of cost and performance to demonstrate \texttt{TurkishBERTweet} is scalable and cost-effective solution.  
As part of our research, we released \texttt{TurkishBERTweet} and fine-tuned LoRA adapters for the mentioned tasks under the MIT License to facilitate future research and
applications on Turkish social media. 
Our \texttt{TurkishBERTweet} model is available at: \texttt{https://github.com/ViralLab/TurkishBERTweet}.

\end{abstract}

\section*{Introduction}

Social media platforms such as Twitter/X have become the primary outlet for individuals to share their opinions on various issues and react to content created by others. 
Increasing use of social media presents an exciting opportunity for researchers to identify trends and analyze online communities shaped by real-world events or organized for common cause~\cite{segerberg2011social,harlow2012social,ogan2017gained,bas2022role}.
However, the informal and concise nature of social media posts can pose challenges for analysis since most models to study textual data trained on formal documents~\cite{baldwin2013noisy,farzindar2015natural}.
Furthermore, the global nature of these platforms introduces an additional layer of complexity with multiple languages being utilized. 
According to \texttt{ethnologue}\footnote{\href{https://www.ethnologue.com/country/TR/}{https://www.ethnologue.com/country/TR/}}, at least 85 million people speak and write Turkish, and Turkish is among the top 20 living languages in the world. 
In 2020, Turkish was ranked as the $11^{th}$ most used language on the Twitter \cite{alshaabi2021growing}, highlighting the importance of research on this widely used language. 
However, it is one of the low-resource languages that lacks annotated datasets for different tasks in NLP \cite{alecakir2022turkishdelightnlp}. 

The recent advances in natural language processing (NLP) let researchers to investigate social media platforms and conduct tasks like sentiment detection, topic modeling, and stance detection more accurately and consistently than traditional approaches. 
There has been a significant improvement in various NLP tasks with the introduction of BERT \cite{devlin-etal-2019-bert}, whose structure is based on the Transformers model \cite{vaswani2017attention}. 
RoBERTa \cite{liu2019roberta} demonstrated that BERT was under-trained and that masked-language modeling would suffice to capture the bidirectional representations of the input. They also utilized Byte-Pair Encoding (BPE) \cite{sennrich2015neural} to encode input texts, which allows the model to learn representation for sub-words, mitigating the out-of-vocabulary (OOV) problem when using the models in an out-of-distribution context. Later, different variants of BERT were introduced to address the need on domain specific datasets. BERTweet \cite{nguyen2020bertweet} is an example of these variants, completely trained on English Twitter datasets.

Models trained with multilingual data also perform better on languages with more training data or data gathering steps for such models tend to have more data quality issues on low-resource languages. 
However, for Turkish Language, the language models that have been developed for Turkish alone are few as it is one of the low-resource languages \cite{alecakir2022turkishdelightnlp}. 
BERTurk \cite{stefan_schweter_2020_3770924}, which is trained on Turkish OSCAR corpus and Wikipedia Dump, is the most popular model that has been employed vastly by Turkish NLP community for wide range of tasks. 
Recently Kesgin \textit{et al.} presented results on transformer-based models trained and evaluated with different model sizes on downstream tasks; however, their contributions were not specifically on a specific domains like social media \cite{toprak2023developing}. 
Recently, foundational models like LLama2 \cite{touvron2023llama} became available open source. These large language models (LLMs) are trained on massive multilingual datasets using significant resources and compute power. Although this particular LLM is not trained with Turkish text, we fine-tune it on Turkish dataset and evaluate its performance. 

In this work, we present TurkishBERTweet, a pre-trained model on Turkish Twitter dataset that contains over 894M tweets spanning several years of online activities to specifically capture the nuanced language used on social media platforms. 
By using TurkishBERTweet, researchers can tackle social media analysis tasks since these platforms contain informal language with irregular vocabularies. 
Combining models like TurkishBERTweet and publicly available social media datasets like \#Secim2023 by our team \cite{najafi2022secim2023}, research community can conduct interdisciplinary research and pursue important societal questions using online data.
We also hope that the Turkish NLP community adopts this model as a strong baseline for their further studies.

Our efforts to develop TurkishBERTweet, a special model for Turkish social media analysis, we made the following contributions:
\begin{itemize}
    \item We introduce the first large-scale pre-trained language model built on large-scale collection of Turkish tweets. We compare this model against different existing models, multi-lingual models, \texttt{Llama-2-7b-hf}, and fine-tuned ChatGPT models.
    \item Our experimental results yield comparable performance to larger pre-trained models (BERTurk) and achieves significantly better results than strong baselines (mBERT and loodos-Albert). TurkishBERTweet also outperforms the SOTA models on Sentiment Analysis and Hate Speech classification tasks.
    \item TurkishBERTweet is the best performing model that can generalize when we experiment with leave-one-dataset-out tasks.
    \item We made our model TurkishBERTweet and LoRA adaptors publicly accessible on Huggingface which can be used with \textit{transformers} library \cite{wolf-etal-2020-transformers}. The codes and experimental results are available on this Github repository: \texttt{https://github.com/ViralLab/TurkishBERTweet}. 
\end{itemize}

\section{TurkishBERTweet}
This section describes the special tokenizer for social media datasets, architecture of our model, Turkish Twitter dataset incorporated for pre-training, and experimental setup we used in this study.

\subsection*{Architecture}
The architecture of are model follows the same structure of the RoBERTa\textsubscript{base} model except the input length of our model is $128$ which makes our model approximately $21.5\textrm{M}$ parameters smaller than BERTurk language model that mimic RoBERTa\textsubscript{base}.


\subsection*{Pre-training Data}
The dataset captures over 10 years of online activity from the Twitter streaming API, covering various important social events and daily discussions. Since the data stream also captures retweeted content, we filtered the retweeted information and retained only the original content posted on the platform.
Our Turkish pre-training dataset includes 110 GB of uncompressed text with nearly 894 million tweets, with each tweet containing at least 10 tokens. Our dataset presents the characteristics of social media posts where few social media accounts responsible with creation of several content and most of the content produced to communicate with other platform users tend to be short texts.



\subsection*{Tokenizer}
Since social media posts have specialized entities, we defined additional tokens to capture them in the documents.
We added extra especial tokens which are  \texttt{@user}, \texttt{\textless{}hashtag\textgreater}, \texttt{\textless{}/hashtag\textgreater}, \texttt{\textless{}cashtag\textgreater}, \texttt{\textless{}/cashtag\textgreater},  \texttt{\textless{}emoji\textgreater}, \texttt{\textless{}/emoji\textgreater},  \texttt{\textless{}http\textgreater}, \texttt{\textless{}/http\textgreater}. The closing tags are for defining the boundaries of especial tokens that will be used for unmasking entities such as emoji, hashtag, etc. 
Having these additional especial tokens in the tokenizer will enable us to prompt out these information through the tweets without any need to do direct supervised learning.

\begin{figure}[!hbt]
    \centering
    \includegraphics[width=\columnwidth]{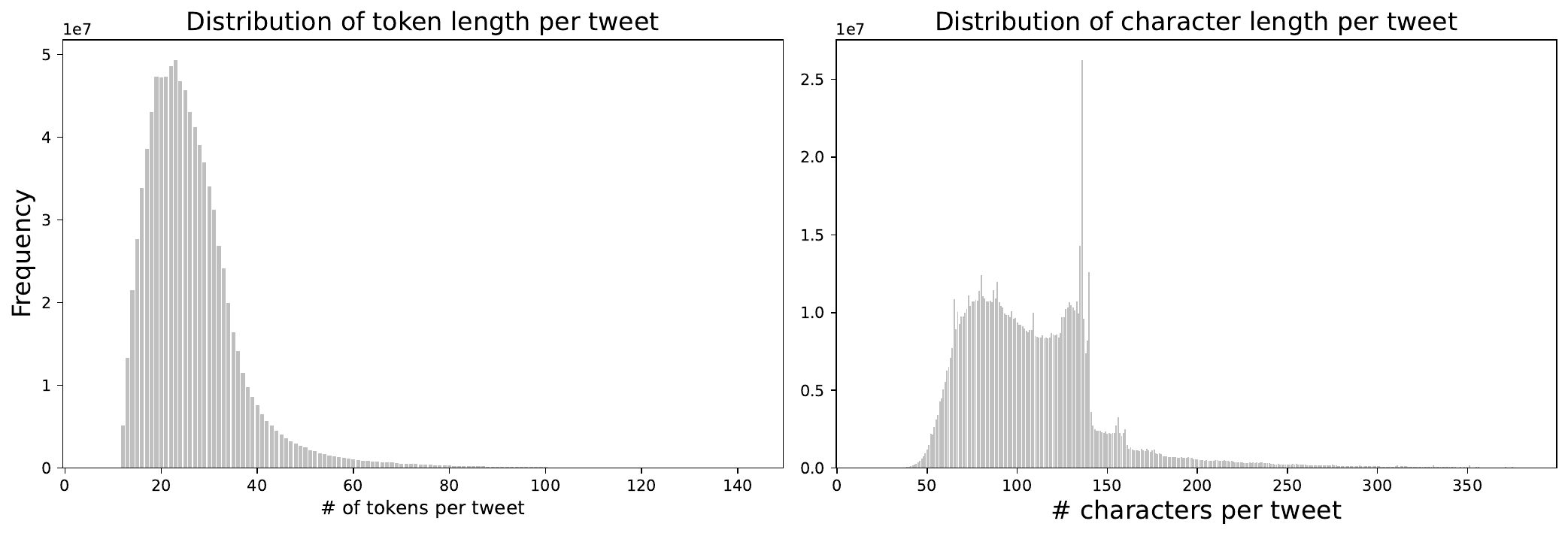}
    \caption{Distribution of the number of tokens per tweet (left) and the distribution of the character length (right) in the pre-trained dataset after tokenization.}
    \label{fig:dist_tokens_chars}
\end{figure}

\begin{figure}[!hbt]
    \centering
    \includegraphics[width=\columnwidth]{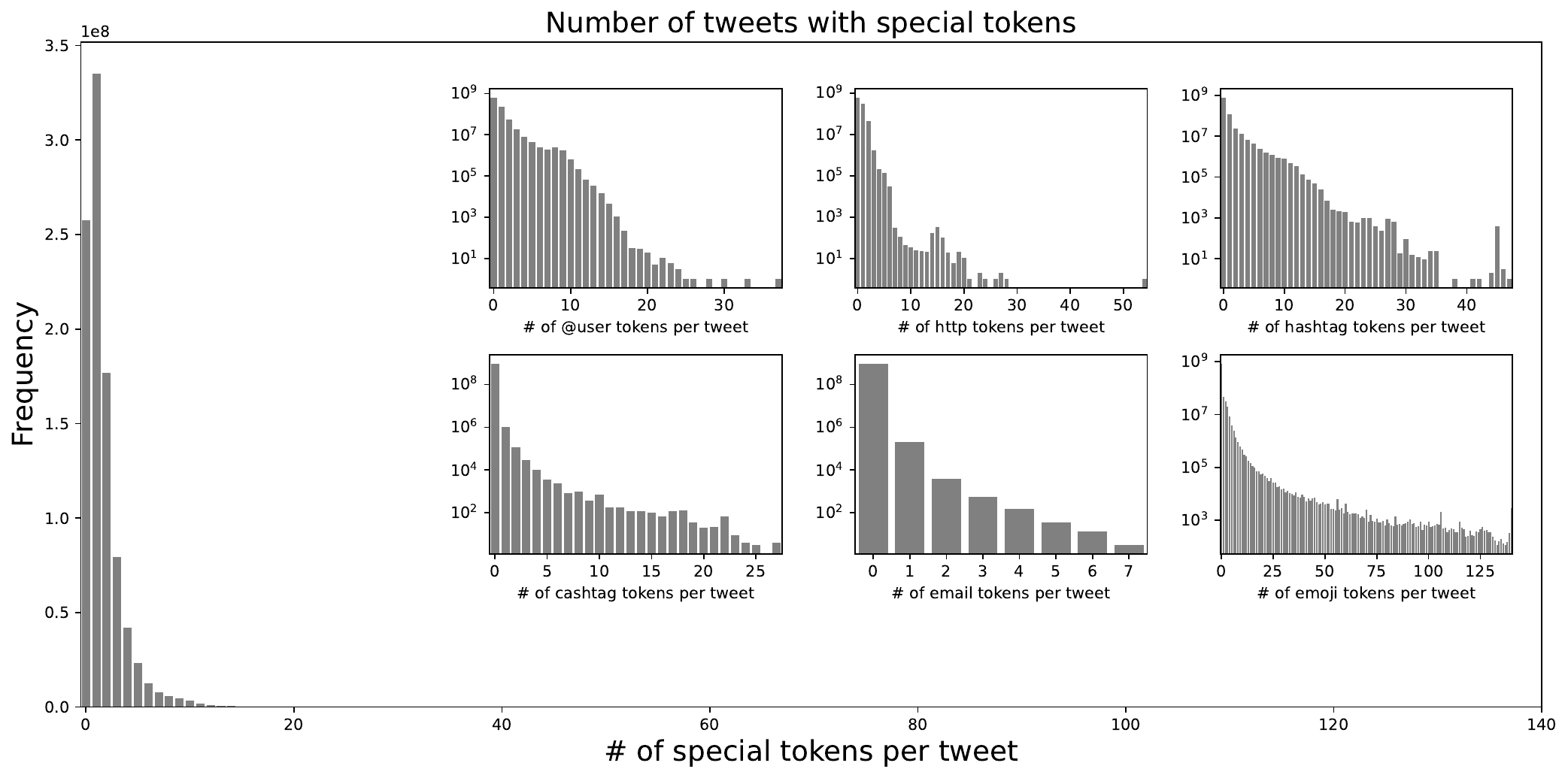}
    \caption{Special Tokens distributions in the pretrained dataset}
    \label{fig:special_token_dist}
\end{figure}

We trained fastBPE \cite{sennrich-etal-2016-neural} tokenizer with vocabulary size of $100,000$. 
Before we fed the data into the model, we applied the preprocessing steps in Table \ref{tab:preprocess_1} to the data. 
We used the \texttt{emoji}\footnote{\url{https://pypi.org/project/emoji}} package to replace the emojis with their equivalent texts, but Turkish equivalents were not available for all emojis. To solve this problem, we used Google Translator to translate them into Turkish. Also, the domains of the url links are extracted and included in the http tokens. In addition, \textbf{@user} token was used instead of mentions and emails with \textbf{\textless{}email\textgreater} in the tweets in order to ensure privacy. By detecting the cash signs in the tweet, we wrapped them with \textbf{cashtag} token. Figure \ref{fig:dist_tokens_chars} shows the distribution of tokens and characters per tweet after preprocessing steps. Figure \ref{fig:special_token_dist} demonstrates the distributions of all of the especial tokens per tweet after preprocessing.

\begin{table}[!htbp]
\begin{adjustbox}{width=0.6\columnwidth,center}
\centering
\begin{tabular}{|c|c|c|}
\hline
\textbf{Entity}& \textbf{Text} & \textbf{Replaced Text}                                                                         \\ \hline
Mention & @foo         & @user                                                                    \\ \hline
Hashtag & \#foo        & \textless{}hashtag\textgreater\hspace{0.1em} foo \textless{}/hashtag\textgreater{}     \\ \hline
Cashtag & \$foo        & \textless{}cashtag\textgreater\hspace{0.1em} foo \textless{}/cashtag\textgreater{}     \\ \hline
Emoji & \emojihuggingface & \textless{}emoji\textgreater\hspace{0.1em} hugging\_face \textless{}/emoji\textgreater{} \\ \hline
URL & www.foo.com & \textless{}http\textgreater\hspace{0.1em} foo \textless{}/http\textgreater{}        \\ \hline
email & info@foo.com & \textless{}email\textgreater{}                                           \\ \hline

\end{tabular}
\end{adjustbox}
\caption{Demonstration of how especial tokens are used to preprocess different entities used in tweet texts.}
\label{tab:preprocess_1}
\end{table}

\subsection*{Optimization}
We utilize the RoBERTa implementation of \texttt{transformers} library. We set the maximum input length as 128 for the model. As a result, (110GB) \* (avg\_n\_subwords) / 128 sequence blocks are generated. 
For optimizing the model, we followed \citet{liu2019roberta} and used Adam (\citet{kingma2014adam}) with the batch size of 128 per each TPU pod, $8 * 128 = 1024$ in total. 
We trained the model for seven days with the peak learning rate of $1\mathrm{e}{-5}$. 

\section{Experimental Setup}

\subsection*{Datasets for Downstream Tasks}
As mentioned earlier, Turkish is one of the low-resource languages for which there are not many annotated data sets. With this in mind, we evaluated the models on two text classification tasks where we can find reliable and sufficient data: Sentiment Analysis and HateSpeech detection. To quantify the consistency and generalizability of the models on novel datasets, we measure the performance of the models not only cross-validated setting but also experimented with the out-of-dataset configuration.

\subsubsection*{Sentiment Analysis}
We evaluate the models on the Sentiment Analysis task having as shown in Table \ref{tab:sa_datasets_stat}. The datasets we used for the experiments and the distribution of their labels presented in the table. It is worth noting that all datasets contain three classes since social media text might be neutral if the discussion is not polarized or simply stating factual information. 

\begin{table}[!htbp]
\begin{adjustbox}{width=1.0\columnwidth,center}
\fontsize{7pt}{7pt}\selectfont

\centering
\begin{tabular}{|lcrrrr|}
\hline
\multicolumn{1}{|l|}{\textbf{Dataset name}}                      & \multicolumn{2}{r|}{\textbf{\# of instance}}       & \multicolumn{1}{r}{\textbf{Positive}}& \multicolumn{1}{r}{\textbf{Neutral}} & \multicolumn{1}{r|}{\textbf{Negative}}     \\ \hline
\multicolumn{1}{|l|}{VRLSentiment}                         & \multicolumn{2}{r|}{23,689}                &   5,469    &  10,146          & 8,074       \\ \hline
\multicolumn{1}{|l|}{TSATweets\tablefootnote{\url{https://github.com/sercankulcu/sentiment-analysis-of-tweets-in-Turkish}}}                    & \multicolumn{2}{r|}{6,001}                     &  1,552    &   1,448   &    3,001      \\ \hline
\multicolumn{1}{|l|}{Kemik-17bin\tablefootnote{\url{http://www.kemik.yildiz.edu.tr/veri_kumelerimiz.html}}}                           & \multicolumn{2}{r|}{17,289}                &   4,579     &      5,822      &  6,888     \\ \hline
\multicolumn{1}{|l|}{Kemik-3000\tablefootnote{\url{http://www.kemik.yildiz.edu.tr/veri_kumelerimiz.html}}}                           & \multicolumn{2}{r|}{3,000}                 &   756      &      957       &  1,287     \\ \hline
\multicolumn{1}{|l|}{BOUN \citep{BounTi}}                & \multicolumn{2}{r|}{4,733}                 &   1,271     &      2,769      &   693     \\ \hline
\multicolumn{1}{|l|}{TSAD\tablefootnote{\url{https://huggingface.co/datasets/winvoker/turkish-sentiment-analysis-dataset}}} & \multicolumn{2}{r|}{489,644}                  &  262,166   & 170,917   &        56,561        \\ \hline
\end{tabular}
\end{adjustbox}
\caption{Descriptive statistics of the datasets and their class distributions.}
\label{tab:sa_datasets_stat}
\end{table}

Researchers have been developing models for Turkish sentiment detection task using deep-learning approaches \cite{ayata2017turkish,ciftci2018deep}.
We search for different publicly available and manually labeled tweet datasets for our experiments. Some datasets provide unique identifiers of tweets, but unfortunately majority of these tweets were either removed or posted by deleted accounts. 
VRLSentiment dataset contains political tweets annotated by students as part of a research project in our group.
We found TSATweets on a Github repository and Kemik datasets are requested from a researcher through email.
The BOUN dataset contains mostly tweets commenting about the universities in Turkey, which means that it covers only a narrow distribution of Twitter platform. 
TSAD dataset differ from other datasets since their source capture product reviews and Turkish Wikipedia.

\subsubsection*{Hate Speech Detection}
We test our model on a hate speech dataset created as part of the Computational Social Sciences Session of the 2023 Signal Processing and Communication Applications Conference (SIU) \cite{arin2023siu2023}.
Organizers released the tweet IDs and their corresponding hate speech classification for the competition. 
We rehydrated all tweets accessible when the dataset was released for the competition. 
Table \ref{tab:hatespeech_stat} shows the distribution of labels in this dataset for binary classification. 
We experiment with the train/test split provided for the evaluation to compare our model with the leaderboard. In addition, we performed a 5-fold cross-validation experiment by combining the training and test sets.

\begin{table}[!htbp]
\centering
\begin{tabular}{lrr} 
\hline
\textbf{Class} & \textbf{Train Set}  & \textbf{Test Set}\\ 
\hline

        No Hate speech    & $3\textrm{,}493$ & $873$    \\
        Hate speech       & $1\textrm{,}190$  & $298$   \\  \hline
        \textbf{Total}    & $\bm{4\textrm{,}683}$  & $\bm{1\textrm{,}171}$ \\\hline
 
\hline
\end{tabular}
\caption{The distribution of classes in Hate Speech dataset.}
\label{tab:hatespeech_stat}
\end{table}


\subsection*{Baselines models for benchmark}
We compare our model with language models, which have various base architectures and are widely used in different fields. These language models are listed below.

\textbf{BERTurk\footnote{\myref{https://huggingface.co/dbmdz/bert-base-turkish-128k-uncased}}:} 
BERTurk is a well-known language model in the Turkish community and has been widely used.
We chose a model with a vocabulary of 128k trained on 35 GB of Turkish text data since this model exists in different versions.
According to the model card on the HuggingFace platform, the model was trained on a collection of OSCAR corpus, a Wikipedia dump, and various OPUS corpora \cite{stefan_schweter_2020_3770924}. 
In OSCAR dataset contains 5,000 tweets \cite{ccarik2022twitter} meaning that the model has seen social media text \cite{2022arXiv220106642A,AbadjiOrtizSuarezRomaryetal.2021,caswell-etal-2021-quality,ortiz-suarez-etal-2020-monolingual,OrtizSuarezSagotRomary2019}. 

\textbf{mBERT\footnote{\myref{https://huggingface.co/bert-base-multilingual-cased}}:} 
This model is trained with the content of the largest 104 languages on Wikipedia. They use a word piece tokenizer and set the vocabulary size to 110k. The Languages with more Wikipedia pages were under-sampled, and the languages with fewer pages were over-sampled to create an input dataset. Unfortunately, no further information is provided regarding the proportion of languages \cite{DBLP:journals/corr/abs-1810-04805}. 

\textbf{mT5-Large\footnote{\myref{https://huggingface.co/google/mt5-large}}:} mT5 is the multilingual version of the T5 language model introduced by (\citeauthor{raffel2020exploring}, \citep{raffel2020exploring}). This model was trained on mC4 datasets that contained almost 71B Turkish tokens, which is 1.93\% of their training dataset \cite{xue2020mt5}.

\textbf{Turkish ALBERT-Base\footnote{\myref{https://huggingface.co/loodos/albert-base-turkish-uncased}}:}	
This model has almost 12M parameters, making it smaller than all other models. It was trained on 200 GB of Turkish text collected from online blogs, free e-books, newspapers, common crawl corpus, Twitter, articles, and Wikipedia. The tokenizer of this model has a 32k vocabulary size. 

\textbf{LLama-2-7b-hf\footnote{\myref{https://huggingface.co/meta-llama/Llama-2-7b-hf}}:} 
Based on the paper of LLama2 \cite{touvron2023llama}, there is no Turkish text in their pre-training dataset, and the majority of their corpus contains English texts. Nevertheless, since this model is a foundation model, we can use it for fine-tuning to see its performance on Turkish texts. We use the 7B version of LLama-2 because we can perform fine-tuning and inference with the computational resources available.

\textbf{GPT3.5-turbo:} Unfortunately, there is no confirmed public information about the training dataset and the pipelines OpenAI used to prepare this model \cite{chatgpt}. We used the paid API of the OpenAI to fine-tune and collect responses for our prompts.


\subsection*{Fine-tuning pre-trained language models}
The fine-tuning procedure uses a pre-trained language model and adopts it for the use in a special task, and there are different methods to build these task-specific models. 
In this work, we experiment with the full fine-tuning and low-rank adaptation (LoRA) fine-tuning methods to compare and evaluate the models for the downstream tasks \cite{hu2021lora}.
To obtain comparable results, we performed 5-fold stratified cross-validation to ensure i) the proportions of the classes are preserved in training and testing and ii) the performance of the models is consistent in each data set. We performed two different fine-tuning approaches, we only conduct experiments for LoRA fine-tuning for the models performed best in the standart fine tuning experiments.

\textbf{Full Fine-tuning (FT)}: In this approach, all or some of the original parameters of the model are updated based on a given data set. Using this approach, we compare our LM with the baselines by freezing all models' parameters and adding a final pooling layer following a dense classification layer. Then, all the models were trained for 50 epochs per fold, and chose the best model for final evaluation of that fold. Early stopping was also used to avoid over-fitting. We used this approach for all of the models except for LLama-2.

\textbf{LoRA Fine-tuning (LFT)}: LoRA is a low-rank adaptation technique for large language models. It works by freezing the pre-trained weights and injecting trainable rank decomposition matrices into each layer of the Transformer architecture. This reduces the number of trainable parameters while still preserving the knowledge learned from the pre-trained model. This approach makes it possible to fine-tune large language models for downstream tasks more efficiently. 
Using the PEFT \cite{peft} library provided by HuggingFace, the models were trained for ten epochs with the rank of $r=8$ and scaling parameter of $alpha = 16$. For the models other than LLama-2, the \texttt{query} and \texttt{value} modules were targeted with the sequence classification objective. For LLama-2, with the same rank and alpha, the model was trained with the context size of $512$ on the Causal Language Modeling (CLM) objective. The model was quantized in 4-bits to prepare for fine-tuning. Since the objective of LLama-2 is CLM, we treat classification as a question-answering task. As a result, we prepare prompts in which the question has the Turkish text embedded after \texttt{Q:} token, and the ground truth label comes after \texttt{A:} token. Table \ref{tab:prompt} shows the prompt structures for the tasks. Similar to the full fine-tuning approach, early stopping was used.

\begin{table}[!htbp]
    \centering
    \begin{adjustbox}{width=\columnwidth,center}
    \begin{tabular}{|c|c|}
    \hline
        Task & Prompt\\    \hline
        Sentiment Analysis & Q: What is the sentiment of this Turkish text: "TEXT"? A: LABEL \\  \hline
        Hate Speech & Q: Does this Turkish text contain hate speech: "TEXT"? A: LABEL\\
    \hline
    \end{tabular}
    \end{adjustbox}
    \caption{Prompt structures for LLama-2-7b}
    \label{tab:prompt}
\end{table}

\textbf{ChatGPT}: 
To finetune the ChatGPT, we prepare the data as the pipelines of OpenAI suggests. Providing contents for three roles of chatbot which are \texttt{system}, \texttt{assistant}, and \texttt{content}. Table \ref{tab:chatgpt} illustrates how the input contents of these roles are structured. We trained on out of distribution datasets for one epoch.

\begin{table}[!htbp]
    \centering
    \begin{adjustbox}{width=\columnwidth,center}
    \begin{tabular}{|c|c|}
    \hline
        Role      & Content \\ \hline
        system    & Vrl-gpt3.5-turbo is a chatbot that can give the sentiment of Turkish texts. \\    \hline 
        user      & What is the sentiment of this Turkish text "TEXT"? \\    \hline 
        assistant & LABEL \\     
    \hline
    \end{tabular}
    \end{adjustbox}
    \caption{Prompt structors for finetuning ChatGPT}
    \label{tab:chatgpt}
\end{table}

\section{Experimental results}

To compare TurkishBERTweet with other available models, we conduct a series of experiments on different datasets we introduced earlier, and we use 5-fold cross-validation for each task. 
Table \ref{tab:sa_hs_RESULTS} presents the results obtained for sentiment analysis and Hate Speech Detection tasks. 
First, we observed significant improvement on both tasks and across different datasets when fine-tuning with LoRA applied in training. 
The two most successful models, BERTurk and TurkishBERTweet, have comparable performance across different datasets for sentiment analysis tasks. 
Since most applications of BERTurk use a standard fine-tuning approach, our publicly available model (TurkishBERTweet model with LoRA) is much preferable and achieves significantly better performance than the BERTurk model with standard fine-tuning.
For 4bit-quantized LLama2-7b, we see lower performance compared to our model and other baselines, but this result was expected since the base model lacks Turkish training data. 

For Hate Speech detection, like sentiment analysis, we performed 5-fold cross-validation to evaluate the performance of the models. We also used the training and testing splits from the SIU 2023 hate speech detection competition. 
Using the dataset set in the competition, we obtained a macro-F1 score of $\bm{0.73167}$ for TurkishBERTweet with LoRA fine-tuning, which is higher than the submission top-ranked in the competition with its score of $0.72167$. 
These scores are reported on the contest page on Kaggle.\footnote{\href{https://www.kaggle.com/competitions/siu2023-nst-task2}{https://www.kaggle.com/competitions/siu2023-nst-task2}}

\begin{figure}[!hbt]
    \centering
    \includegraphics[width=0.5\columnwidth]{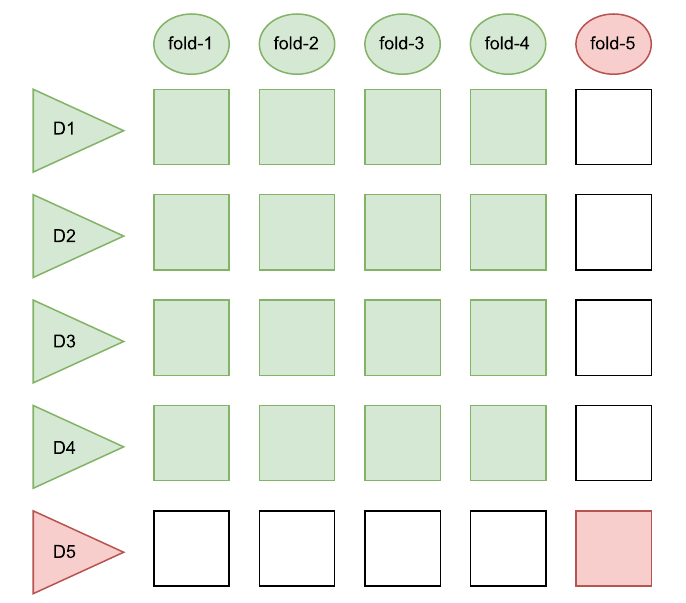}
    \caption{5-Fold Cross Validation across different datasets. The green and red colors represent train and test splits for the first fold, respectively.}
    \label{fig:OOD}
\end{figure}

To investigate the generalizability of the models on different domains, we performed an out-of-distribution evaluation in which we left one of the datasets out and trained the models on the rest of the datasets. 
We selected the test fold of the test dataset to compare the results obtained in this configuration with the single dataset configuration. We grouped the training folds of the remaining datasets. Figure \ref{fig:OOD} illustrates the split of training and test for the OOD evaluation. We selected the best models to observe their performance in this setting, which are TurkishBERTweet+LoRA and BERTurk+LoRA.
We witnessed a decrease in performance for both TurkishBERTweet+LoRA and BERTurk, as expected since the testing datasets are different from the ones provided for training. 
It is worth mentioning that TurkishBERTweet+LoRA still outperforms the BERTurk language model almost on all of the datasets except BOUN. Besides, we fine-tuned the ChatGPT3.5 Turbo model in this setting, and it performed almost equally compared to the competitors.

\newcommand{\STAB}[1]{\begin{tabular}{@{}c@{}}#1\end{tabular}}
{\def\arraystretch{1.3}
\begin{table}[!t]
  \centering
  \begin{adjustbox}{width=1.4\columnwidth,center}
  \begin{tabular}{|c|c|c|c|c|c|c|c|c||c|}
    \hline
      &\multicolumn{1}{c|}{Task} & \multicolumn{1}{c|}{Model / Dataset} &\multicolumn{1}{c|}{VRLSentiment} & \multicolumn{1}{c|}{Kemik-17bin} & \multicolumn{1}{c|}{Kemik-3000} & \multicolumn{1}{c|}{TSATweets} & \multicolumn{1}{c|}{BOUN} & \multicolumn{1}{c||}{TSAD}  & \multicolumn{1}{c|}{HateSpeech}\\ \hline
     \multirow{3}{*}{\STAB{\rotatebox[origin=c]{90}{\textbf{LFT}}}}
     & SEQ\_CLS & TurkishBERTweet&$\bm{0.635\pm0.007}$  & $\bm{0.757\pm0.008}$ & $\bm{0.661\pm0.014}$ & $\bm{0.710\pm0.010}$ & $\bm{0.729\pm0.014}$ & $0.968\pm0.001$ & $\bm{0.808\pm0.011}$ \\
     & SEQ\_CLS &BERTurk& $\bm{0.638\pm0.007}$ & $\bm{0.771\pm0.007}$ & $\bm{0.681\pm0.027}$ & $\bm{0.706\pm0.010}$ & $\bm{0.752\pm0.018}$ & $\bm{0.972\pm0.001}$ & $\bm{0.812\pm0.009}$ \\
     & SEQ2SEQ &google/mt5-large&$0.631\pm0.007$ & $0.713\pm0.012$ & $0.524\pm0.043$ & $0.686\pm0.016$ & $0.726\pm0.029$ & $\bm{0.975\pm0.001}$  & $0.755\pm0.015$ \\
     & SEQ2SEQ & LLama2-7b-hf& $0.571\pm0.032$ & $0.675\pm0.009$ & $0.573\pm0.025$ & $0.563\pm0.068$ & $0.643\pm0.023$ & NA & $0.703\pm0.021$ \\\hline
     \multirow{4}{*}{\STAB{\rotatebox[origin=c]{90}{\centering{\textbf{FT}}}}}
     & SEQ\_CLS &TurkishBERTweet&$\bm{0.611\pm0.007}$ & $\bm{0.688\pm0.006}$ & \bm{$0.602\pm0.010$} & \bm{${0.638\pm0.009}$} & $0.644\pm0.012$ & $0.898\pm0.001$ &  \bm{$0.729\pm0.006$} \\
     & SEQ\_CLS &BERTurk&$0.598\pm0.006$ & \bm{$0.688\pm0.008$} & \bm{$0.626\pm0.012$} & \bm{$0.644\pm0.012$} & \bm{$0.691\pm0.022$} & \bm{$0.932\pm0.001$} & \bm{$0.723\pm0.013$} \\
     &SEQ\_CLS & mBERT& $0.539\pm0.003$ & $0.586\pm0.010$ & $0.508\pm0.010$ & $0.576\pm0.013$ & $0.628\pm0.031$ & $0.859\pm0.001$ & $0.689\pm0.017$ \\
     & SEQ\_CLS &Turkish Albert-BASE&$0.552\pm0.003$ & $0.641\pm0.006$ & $0.572\pm0.014$ & $0.607\pm0.010$  & $0.644\pm0.019$ & $0.881\pm0.001$ & $0.715\pm0.009$ \\\hline
     \multirow{1}{*}{\STAB{\rotatebox[origin=c]{90}{\centering{\textbf{ZS}}}}}
     & SEQ2SEQ &LLama2-7b-hf&$0.437$ & $0.454$ & $0.458$ & $0.455$ & $0.434$ &  NA & $0.4420$ \\
     
     \hline
     \end{tabular}
    \end{adjustbox}
    \caption{Weighted F1-score of the baseline models for Sentiment Analysis and Hate Speech Tasks. Best scores are presented in bold font and when the difference is not significant more than one model highlighted.}
     \label{tab:sa_hs_RESULTS}
\end{table} 
}

\begin{table}[!htbp]
\begin{adjustbox}{width=\columnwidth,center}
\centering
\fontsize{7pt}{7pt}\selectfont

\begin{tabular*}{0.85\linewidth}{ p{0.175\textwidth} |c|c|c} 
\hline
\multicolumn{1}{c|}{\textbf{Experiment}} & \multicolumn{1}{c|}{\textbf{TurkishBERTweet+LoRA}} & \multicolumn{1}{c|}{\textbf{BERTurk}} & \multicolumn{1}{c}{\textbf{gpt3.5-turbo}} \\ \hline

$\forall-\textrm{VRLSentiment}$      & $\bm{0.546\pm0.004}$ & $0.530\pm0.003$ & $\bm{0.5551}$ \\ 
$\forall-\textrm{Kemik--17bin}$    & $\bm{0.638\pm0.011}$ & $0.634\pm0.006$  &  $\bm{0.6527}$ \\ 
$\forall-\textrm{Kemik--3000}$     & $\bm{0.634\pm0.016}$ & $0.603\pm0.019$  & $\bm{0.6366}$ \\ 
$\forall-\textrm{TSATweets}$       & $\bm{0.594\pm0.007}$ & $0.510\pm0.015$  & $0.5493$ \\ 
$\forall-\textrm{BOUN}$            & $0.599\pm0.011$ & $\bm{0.667\pm0.013}$  & $0.5802$\\
\hline
\end{tabular*}
\end{adjustbox}
\caption{Weighted F1-score for leave one dataset out evaluation. In each experiment one dataset used for evaluation while others are used in model training.}
\label{tab:oov_dataset_results}
\end{table}

In additional to compare models based on performance, we can also measure inference time and models size to consider their usability in large-scale analysis. 
In terms of input length of the models, TurkishBERTweet works with input length of 128, which is half of the input length for BERTurk. 
This property of the model reduces the size of the model. 
Consequently, the batch size can be increased to load more data onto the GPU during inference time. 
To compare the performance of the models, we ran a sample of 1,000 Turkish tweets with the maximum batch size that can be used for the models to capture the best inference time. 
It should be mentioned that we padded the input texts to the maximum input length of the models. 
For the experiment, we used \texttt{1X NVIDIA GeForce RTX 4090 24GB}. We repeated the experiment for each model 100 times and report the average inference time in Table \ref{tab:inference_eval}. 
This practical comparison points that TurkishBERTweet model is more suitable to process millions of tweets in significantly faster for social media analysis. For instance, firehose data stream (all public tweet) of Twitter produce about 4,000 tweets per second \cite{pfeffer2023just}. Considering less than 10\% of public tweets posted in Turkish, we can process such data streams in real time.

\begin{table}[!htbp]
\begin{adjustbox}{width=0.80\columnwidth,center}
\centering
\fontsize{7pt}{7pt}\selectfont
\begin{tabular*}{0.60\linewidth}{p{0.175\textwidth}| r | r | c} 
\hline
\multicolumn{1}{l}{\textbf{Model}} & \multicolumn{1}{l}{\textbf{Size}}  & \multicolumn{1}{l}{\textbf{Inference Time}} & \multicolumn{1}{l}{\textbf{Batch Size}}\\ \hline
        TurkishBERTweet        & $163\textrm{M}$ & $1.2244$ secs  & 1024  \\
        TurkishAlbert-Base     & $12\textrm{M}$  & $1.1397$ secs            & 1024  \\  
        BERTurk                & $184\textrm{M}$  & $3.2004$ secs           &  512   \\  
        mBERT                  & $177\textrm{M}$  & $3.2001$ secs           &  512  \\  
        mt5-Large              & $974\textrm{M}$  & $37.2270$ secs          &  64  
        \\  
\hline
\end{tabular*}
\end{adjustbox}
\caption{Model sizes and their inference time on 1000 samples.}
\label{tab:inference_eval}
\end{table}

\section{Discussion}
Building a language model specially trained from Turkish social media posts taught us valuable lessons during the process. 
When we began pre-training TurkishBERTweet on Twitter/X data, we hypothesized that a fully Turkish tweets dataset would improve results on downstream tasks. 
As we showed in evaluation section, our model is achieving almost similar results on tasks with BERTurk, expect the LoRA fine-tuning led to significant performance improvement compared to other publicly available models.
This finding is approximately aligned with the findings in BERTweet paper\cite{nguyen2020bertweet}, which is with a smaller model pre-trained on domain-specific datasets, better performance can be achieved. The authors achieved almost 2 points better F1-score in the text classification and almost the same performance in the NER task. This finding is also consistent with the discussion about the quality of the pre-trained dataset \citet{longpre2023pretrainer}.

We also recall that compared to LLama2 for the downstream tasks, our model performed much better, but it is also fair to mention that quantization during fine-tuning reduces the performance of the original model. It may be possible to achieve higher performance with full fine-tuning of one of these huge models, but at the cost of computational resources. All in all, this is a trade-off for users of these huge LLMs. 
Since we used LLama2 model with 7B parameters, one cannot generalize that these models need to be better for Turkish NLP. It is also good to mention that prompt construction is another factor in the performance of generative models, and their performance can be improved if more context is given about the task. We did not explore this topic in depth because the prompt construction of generative models is outside our research scope. We see the random performance of LLama2 in the Zero Shot classification setting, which was predictable because the model has not seen any Turkish texts.

In the context of OOD evaluation, we see a decrease in the models' performance compared to the single dataset evaluation. This outcome is expected to a certain level since each dataset may have similar instances across training and test sets; however, different datasets can vary temporally and topically. We also observed that the performance of ChatGPT was almost similar to the performance of our proposed model, emphasizing that our model is open-source and free of charge.


In the Turkish NLP community, almost all available sentiment analysis models are binary classification models, meaning that an input is either positive or negative, which is not always the case since a text can also have a neutral sentiment. To fill this gap, we provide our final Sentiment Analysis model as a three-class classifier.

\subsection*{Applications of LLM}
\begin{figure}[!hbt]
    \centering
    \includegraphics[width=\columnwidth]{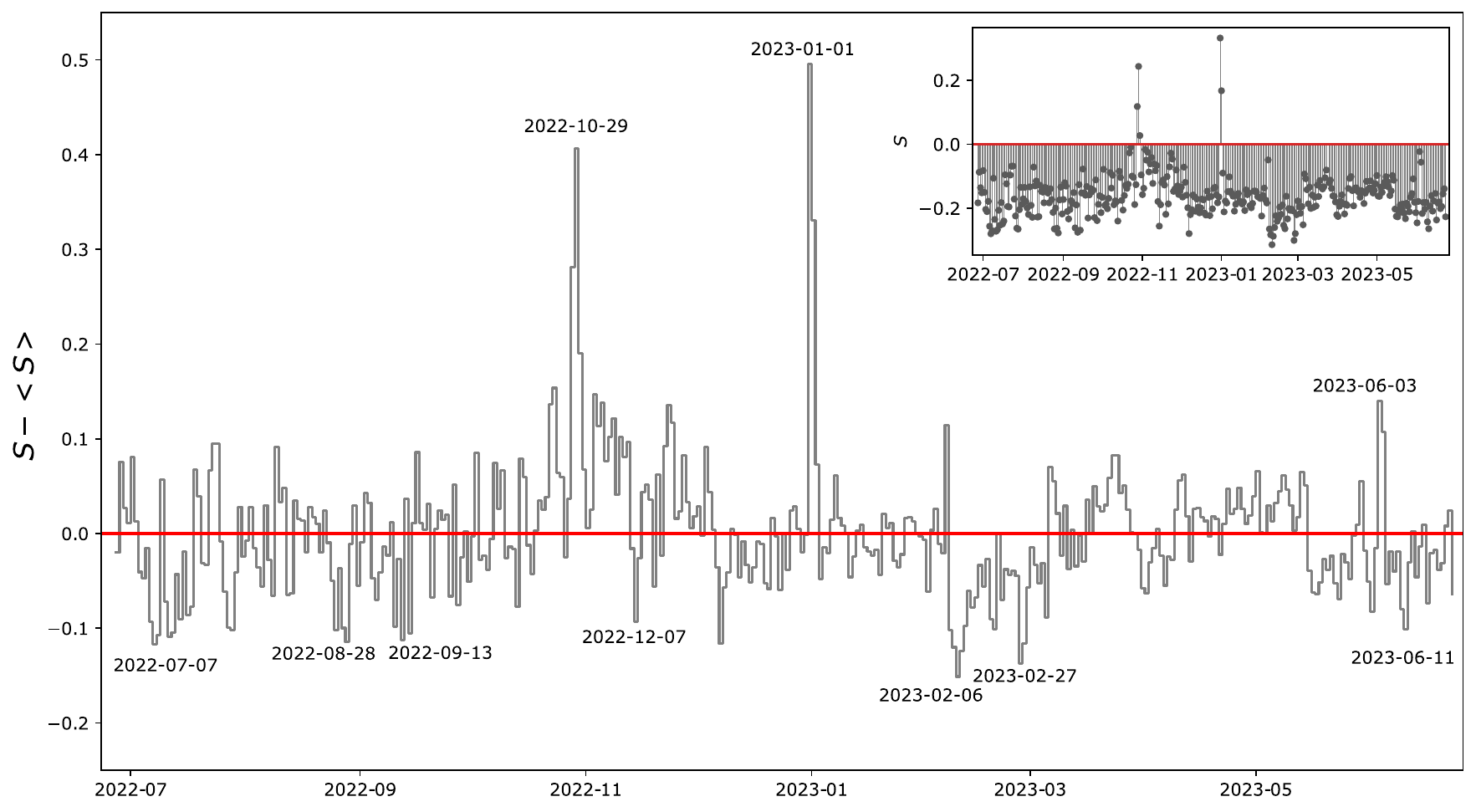}
    \caption{\textbf{Daily sentiment, and sentiment difference from the mean.} $S$ and $\textlangle S \textrangle$ stand for sentiment and mean sentiment of tweets.}
    \label{fig:sentiment_election}
\end{figure}

One of the main use cases of the Languge model, which we have presented in this article, is its use in research projects dealing with large amounts of data. Since TurkishBERTweet is an open source model with better performance and faster inference, it is a good choice for such projects. One of the examples is analyzing the sentiments of the dataset \#Secim2023 \cite{najafi2022secim2023}, which consists of $\bm{336,690,250}$ tweets ranging from July 2021 to June 2023. In Fig.\ref{fig:sentiment_election} presents the daily aggregated sentiment deviances, and the daily aggregated sentiments for the last one year year. The dates of the days with extreme sentiment values are also mentioned in the figure. For instance, Feb 6, 2023, is highly negative as the unfortunate Turkey-Syria earthquake happened.



\subsection*{Cost Estimation}
Based on OpenAI's current pricing policy\footnote{https://openai.com/pricing}, GPT3.5 Turbo for inference costs \$0.001 and \$0.002 per 1,000 tokens for input and output usage, respectively. For the dataset mentioned in the previous section, the number of tokens in the Election dataset using OpenAI's token counter is over 40.2 billion, which means that only inference costs nearly \textbf{\$41K}, which is costly for research usage. The equation \ref{formula_inference} consists of two parts: the input cost and the output cost. The coefficient three results from generating three tokens as output based on the output prompt structure. The final cost of the inference is the sum of the input and output costs.

\begin{equation}
\begin{split}
    &InferenceCost_{Input}(N_{tokens}) = N_{tokens} * 0.000001  \\
    &InferenceCost_{Output}(N_{tweets}) = N_{tweets} * 3 * 0.000002 \\
    &Inference Cost = InferenceCost_{Input} + InferenceCost_{Output}
    \label{formula_inference}
\end{split}
\end{equation}

Considering the extreme budget requirement of commercial models, \texttt{TurkishBERTweet} model as a free alternative offers the same performance. For our experiments with out-of-datasets, we fine-tuned models shown in Table \ref{tab:oov_dataset_results}. The fine-tuning for these models cost about 130\$.

\subsection*{Limitations}
Turkish is one of the most widely used languages on social media platforms. In this study, we attempted to evaluate our model for NLP tasks other than text classification. However, there are no open source Twitter datasets for tasks such as Named Entity Recognition and POS tagging or the available papers online share Tweet IDs \cite{kuccuk2019tweet}, which prevented us from further comparisons with the baseline models. Those that share annotations and Tweet IDs are no longer usable due to deletions and, more importantly, lack of API access to Twitter/X. 
Due to computational limitations, we were unable to increase the context length of our model to capture more context through text. 
Higher quantization bits to fine-tune \texttt{Llama-2-7b-hf} was also not feasible, as this requires high VRAM capacity.
 


\section{Conclusions}
TurkishBERTweet is the first language model pre-trained on nearly 900 million Turkish tweets. 
We present this language model as a SOTA toolkit for the Turkish NLP community and other researchers. We show its better performance on two text classification tasks than the baselines BERTurk, LLama2, and the fine-tuned ChatGPT model. 
TurkishBERTweet is a lightweight model that is computationally much more efficient, so researchers can easily use it for their research tasks. Moreover, we showed that for data-extensive research that needs a significant amount of inferences, ChatGPT is costly. As it is a close source, we need to share our data with OpenAI to use it, which is a downside, especially when dealing with sensitive data. 

\section*{Acknowledgments}
We thank Fatih Amasyali for providing access to Tweet Sentiment datasets from Kemik group.
This material is based upon work supported by the Google Cloud Research Credits program with the award GCP19980904. We also thank TUBITAK (121C220 and 222N311) for funding this project. 


\bibliographystyle{unsrtnat}  
\footnotesize
\bibliography{main}  

\end{document}